\documentclass[wcp]{jmlr}

\usepackage{longtable}% for long tables

\usepackage{booktabs}
\usepackage[T1]{fontenc}
\usepackage{booktabs}
\usepackage{listings}
\usepackage{esvect}
\usepackage{xspace}
\usepackage{algorithm,algorithmic}
\usepackage{mathrsfs}
\usepackage{enumerate}
\usepackage{nicefrac} 
\usepackage{microtype}
\usepackage{makecell}
\usepackage{lscape}
\usepackage{color}
\usepackage{hyperref,cleveref}

\makeatletter
\let\Ginclude@graphics\@org@Ginclude@graphics 
\makeatother
\jmlrvolume{157}
\jmlryear{2021}
\jmlrworkshop{ACML 2021}

\usepackage[nolist]{acronym}
\begin{acronym}[UML]
	\acro{AI}{Artificial Intelligence}
	\acro{BFS}{Breadth-First-Search}
	\acro{BoW}{Bag-of-Words}
	\acro{CSV}{Comma-Separated Values}
	\acro{CNN}{Convolutional Neural Network}
	\acro{DL}{Description logic}

	\acro{ER}{Entity Resolution}
	\acro{EM}{Expectation Maximization}
	\acro{EBMT}{Example-Based Machine Translation}
	\acro{EBNF}{Extended Backus--Naur Form}
	\acro{EL}{Entity Linking}
	\acro{FAO}{Food and Agriculture Organization of the United Nations}
	\acro{GIS}{Geographic Information Systems}
	\acro{GHO}{Global Health Observatory}
	\acro{GRU}{Gated recurrent unit}
	\acro{HDI}{Human Development Index}
	\acro{ICT}{Information and communication technologies}
	\acro{IFRS}{International Financial Reporting Standards}
	\acro{ICD}{International Classification of Diseases}
	\acro{IT}{Information Technology}
    \acro{KB}{Knowledge Base}
    \acro{KG}{Knowledge Graph}
    \acrodefplural{KG}{Knowledge Graphs}
    \acro{KGE}{Knowledge Graph Embeddings}
    \acro{KBSE}{Knowledge Base Semantic Embedding}
    \acro{KGC}{Knowledge Graph Completion}
	\acro{LR}  {Language Resource}
	\acro{LD}  {Linked Data}
	\acro{LLOD}  {Linguistic Linked Open Data}
	\acro{LIMES}{LInk discovery framework for MEtric Spaces}
	\acro{LS}  {Link Specifications}
	\acro{LDIF}{Linked Data Integration Framework}
	\acro{LGD} {LinkedGeoData}
	\acro{LOD} {Linked Open Data}
	\acro{LOV} {Linked Open Vocabularies}
	\acro{LSTM}{Long Short-Term Memories}
	\acro{MSE}{Mean Squared Error}
	\acro{ML}{Machine Learning}
	\acro{MR}{Machine Learning}
	\acro{MRR}{Mean Reciprocal Rank}
	
	\acro{NIF}{Natural Language Processing Interchange Format}
	\acro{NIF4OGGD}{NLP Interchange Format for Open German Governmental Data}
	\acro{NLP}{Natural Language Processing}
	\acro{NER}{Named Entity Recognition}
	\acro{NMT}{Neural Machine Translation}
	\acro{NN}{Neural Network}
	\acrodefplural{NN}{Neural Networks}
	\acro{NLG}{Natural Language Generation}
	\acro{NED}{Named Entity Disambiguation}
	\acro{NERD}{Named Entity Recognition and Disambiguation}
	\acro{NL}{Natural Language}
	\acro{NIF}{NLP Interchange Format}
	\acro{NIF4OGGD}{NLP Interchange Format for Open German Governmental Data}
	\acro{NLP}{Natural Language Processing}
	\acro{NER}{Named Entity Recognition}
	\acro{NEL}{Named Entity Linking}
	\acro{NE}{Named Entity}
	\acro{NN}{Neural Network}
	\acro{NLI}{Natural Language Inference}
	\acro{OSM}{OpenStreetMap}
	\acro{OWL}{Web Ontology Language}
	\acro{OOV}{out-of-vocabulary}
	\acro{PFM}{Pseudo-F-Measures}
	\acro{PSO}{Particle-Swarm Optimization}
	\acro{PBSMT}{Phrase-Based Statistical Machine Translation}
	
	\acro{QA}{Question Answering}
	\acro{RDF}{Resource Description Framework}
	\acro{RBMT}{Rule-Based Machine Translation}
	\acro{RNN}{Recurrent Neural Network}
	\acro{ReLU}{rectified linear unit}
	\acro{RDFS}{RDF Schema}
	\acro{SKOS}{Simple Knowledge Organization System}
	\acro{SPARQL}{SPARQL Protocol and RDF Query Language}
	\acro{SRL}{Statistical Relational Learning}
	\acro{SWT}{Semantic Web Technologies}
	\acro{SW}{Semantic Web}
	\acro{SMT}{Statistical Machine Translation}
	\acro{SWMT}{Semantic Web Machine Translation}
	\acro{SKOS}{Simple Knowledge Organization System}
	\acro{SPARQL}{SPARQL Protocol and RDF Query Language}
	\acro{SRL}{Statistical Relational Learning}
	\acro{SF}{surface forms}
    \acro{TBMT} {Transfer-Based Machine Translation}
	\acro{UML}{Unified Modeling Language}
	\acro{USL}{Ukrainian Sign Language}
	\acro{URI}{Uniform Resource Identifier}
	\acro{WHO}{World Health Organization}
	\acro{WKT}{Well-Known Text}
	\acro{W3C}{World Wide Web Consortium}
	\acro{WSD}{Word Sense Disambiguation}
	\acro{WMT}{Workshop on Machine Translation}
    \acro{XML}{Extensible Markup Language}
	\acro{YPLL}{Years of Potential Life Lost}
	\acro{AOS}{Agricultural Ontology Services}
	\acro{AGRIS}{Agricultural Science and Technology}
	\acro{API}{Application Programming Interface}
	\acro{AOS}{Agricultural Ontology Services}
	\acro{AGRIS}{Agricultural Science and Technology}
	\acro{API}{Application Programming Interface}
	\acro{A2KB}{Annotation to Knowledge Base}
	\acro{BPSO}{Binary Particle-Swarm Optimization}
	\acro{BPMLOD}{Best Practices for Multilingual Linked Open Data}
	\acro{BPSO}{Binary Particle-Swarm Optimization}
	\acro{BPMLOD}{Best Practices for Multilingual Linked Open Data}
	\acro{BFS}{Breadth-First-Search}
	\acro{BPE}{Byte Pair Encoding}
	\acro{BoW}{Bag-of-Words}
	\acro{CBD}{Concise Bounded Description}
	\acro{COG}{Content Oriented Guidelines}
	\acro{CSV}{Comma-Separated Values}
	\acro{CBMT}{Corpus-Based Machine Translation}
	\acro{CLIR}{Cross-Language Information Retrieval}
	\acro{DPSO}{Deterministic Particle-Swarm Optimization}
	\acro{DALY}{Disability Adjusted Life Year}

	\acro{ER}{Entity Resolution}
	\acro{EM}{Expectation Maximization}
	\acro{EBMT}{Example-Based Machine Translation}
	\acro{EBNF}{Extended Backus--Naur Form}
	\acro{EL}{Entity Linking}
	\acro{FAO}{Food and Agriculture Organization of the United Nations}
	\acro{GIS}{Geographic Information Systems}
	\acro{GHO}{Global Health Observatory}
	\acro{GRU}{Gated recurrent unit}
	\acro{HDI}{Human Development Index}
	\acro{ICT}{Information and communication technologies}
	\acro{IFRS}{International Financial Reporting Standards}
	\acro{ICD}{International Classification of Diseases}
	\acro{IT}{Information Technology}
	\acro{IRI}{International Resource Identifier}
    \acro{KB}{Knowledge Base}
    \acro{KG}{Knowledge Graph}
    \acro{KGE}{Knowledge Graph Embeddings}
    \acro{KBSE}{Knowledge Base Semantic Embedding}
	\acro{LR}  {Language Resource}
	\acro{LD}  {Linked Data}
	\acro{LLOD}  {Linguistic Linked Open Data}
	\acro{LIMES}{LInk discovery framework for MEtric Spaces}
	\acro{LS}  {Link Specifications}
	\acro{LDIF}{Linked Data Integration Framework}
	\acro{LGD} {LinkedGeoData}
	\acro{LOD} {Linked Open Data}
	\acro{LSTM}{Long Short-Term Memories}
	\acro{MSE}{Mean Squared Error}
	\acro{MWE}{Multiword Expressions}
	\acro{MT}{Machine Translation}
	\acro{ML}{Machine Learning}
	\acro{MR}{Machine Reading}
	\acro{MOS}{Manchester OWL Syntax}
	\acro{NIF}{Natural Language Processing Interchange Format}
	\acro{NIF4OGGD}{NLP Interchange Format for Open German Governmental Data}
	\acro{NLP}{Natural Language Processing}
	\acro{NER}{Named Entity Recognition}
	\acro{NMT}{Neural Machine Translation}
	\acro{NN}{Neural Network}
	\acro{NLG}{Natural Language Generation}
	\acro{NED}{Named Entity Disambiguation}
	\acro{NERD}{Named Entity Recognition and Disambiguation}
	\acro{NL}{Natural Language}
	\acro{NIF}{NLP Interchange Format}
	\acro{NIF4OGGD}{NLP Interchange Format for Open German Governmental Data}
	\acro{NLP}{Natural Language Processing}
	\acro{NER}{Named Entity Recognition}
	\acro{NEL}{Named Entity Linking}
	\acro{NE}{Named Entity}
	\acro{NN}{Neural Network}
	\acro{NLI}{Natural Language Inference}
	\acro{OSM}{OpenStreetMap}
	\acro{OWL}{Web Ontology Language}
	\acro{OOV}{out-of-vocabulary}
	\acro{PFM}{Pseudo-F-Measures}
	\acro{PSO}{Particle-Swarm Optimization}
	\acro{PBSMT}{Phrase-Based Statistical Machine Translation}
	
	\acro{QA}{Question Answering}
	\acro{RDF}{Resource Description Framework}
	\acro{RBMT}{Rule-Based Machine Translation}
	\acro{RNN}{Recurrent Neural Network}
	\acro{ReLU}{rectified linear unit}
	\acro{SKOS}{Simple Knowledge Organization System}
	\acro{SPARQL}{SPARQL Protocol and RDF Query Language}
	\acro{SRL}{Statistical Relational Learning}
	\acro{SWT}{Semantic Web Technologies}
	\acro{SW}{Semantic Web}
	\acro{SMT}{Statistical Machine Translation}
	\acro{SWMT}{Semantic Web Machine Translation}
	\acro{SKOS}{Simple Knowledge Organization System}
	\acro{SPARQL}{SPARQL Protocol and RDF Query Language}
	\acro{SRL}{Statistical Relational Learning}
	\acro{SF}{surface forms}
	\acro{SVM}{Support Vector Machines}
    \acro{TBMT} {Transfer-Based Machine Translation}
	\acro{UML}{Unified Modeling Language}
	\acro{USL}{Ukrainian Sign Language}
	\acro{URI}{Uniform Resource Identifier}
	\acro{WHO}{World Health Organization}
	\acro{WKT}{Well-Known Text}
	\acro{W3C}{World Wide Web Consortium}
	\acro{WSD}{Word Sense Disambiguation}
	\acro{WWW}{World Wide Web}
    \acro{XML}{Extensible Markup Language}
	\acro{YPLL}{Years of Potential Life Lost}
	\acro{AOS}{Agricultural Ontology Services}
	\acro{AGRIS}{Agricultural Science and Technology}
	\acro{API}{Application Programming Interface}
	\acro{BPSO}{Binary Particle-Swarm Optimization}
	\acro{BPMLOD}{Best Practices for Multilingual Linked Open Data}
	\acro{CBD}{Concise Bounded Description}
	\acro{COG}{Content Oriented Guidelines}
	\acro{CSV}{Comma-Separated Values}
	\acro{CBMT}{Corpus-Based Machine Translation}
	\acro{CLIR}{Cross-Language Information Retrieval}
	\acro{DPSO}{Deterministic Particle-Swarm Optimization}
	\acro{DALY}{Disability Adjusted Life Year}
	\acro{DRL}{Deep Reinforcement Learning}

	\acro{ER}{Entity Resolution}
	\acro{EM}{Expectation Maximization}
	\acro{EBMT}{Example-Based Machine Translation}
	\acro{EBNF}{Extended Backus--Naur Form}
	\acro{EL}{Entity Linking}
	\acro{FAO}{Food and Agriculture Organization of the United Nations}
	\acro{GIS}{Geographic Information Systems}
	\acro{GHO}{Global Health Observatory}
	\acro{HDI}{Human Development Index}
	\acro{ICT}{Information and communication technologies}
    \acro{KB}{Knowledge Base}
    \acro{KBSE}{Knowledge Base Semantic Embedding}
	\acro{LR}  {Language Resource}
	\acro{LD}  {Linked Data}
	\acro{LLOD}  {Linguistic Linked Open Data}
	\acro{LIMES}{LInk discovery framework for MEtric Spaces}
	\acro{LS}  {Link Specifications}
	\acro{LDIF}{Linked Data Integration Framework}
	\acro{LGD} {LinkedGeoData}
	\acro{LOD} {Linked Open Data}
	\acro{ML}{Machine Learning}
	\acro{MDP}{Markov Decision Process}
	\acro{MT}{Machine Translation}
	\acro{NIF}{Natural Language Processing Interchange Format}
	\acro{NIF4OGGD}{NLP Interchange Format for Open German Governmental Data}
	\acro{NLP}{Natural Language Processing}
	\acro{NER}{Named Entity Recognition}
	\acro{NMT}{Neural Machine Translation}
	\acro{NN}{Neural Network}
	\acro{NLG}{Natural Language Generation}
	\acro{NED}{Named Entity Disambiguation}
	\acro{NERD}{Named Entity Recognition and Disambiguation}
	\acro{NL}{Natural Language}
	\acro{OWL}{Web Ontology Language}
	\acro{QA}{Question Answering}
	\acro{RDF}{Resource Description Framework}
	\acro{RBMT}{Ru\-le-Ba\-sed Ma\-chi\-ne Trans\-la\-tion}
	\acro{REG}{Referring Expression Generation}
	\acro{SKOS}{Simple Knowledge Organization System}
	\acro{SPARQL}{SPARQL Protocol and RDF Query Language}
	\acro{SRL}{Statistical Relational Learning}
	\acro{SWT}{Semantic Web Technologies}
	\acro{SW}{Semantic Web}
	\acro{SMT}{Statistical Machine Translation}
	\acro{SWMT}{Semantic Web Machine Translation}
    \acro{TBMT} {Transfer-Based Machine Translation}
	\acro{VS}{Version Space}

	\acro{WHO}{World Health Organization}
	\acro{WKT}{Well-Known Text}
	\acro{W3C}{World Wide Web Consortium}
	\acro{WSD}{Word Sense Disambiguation}
    \acro{XML}{Extensible Markup Language}
	\acro{YPLL}{Years of Potential Life Lost}
	\acro{AOS}{Agricultural Ontology Services}
	\acro{AGRIS}{Agricultural Science and Technology}
	\acro{API}{Application Programming Interface}
	\acro{AOS}{Agricultural Ontology Services}
	\acro{AGRIS}{Agricultural Science and Technology}
	\acro{API}{Application Programming Interface}
	\acro{A2KB}{Annotation to Knowledge Base}
	\acro{BPSO}{Binary Particle-Swarm Optimization}
	\acro{BPMLOD}{Best Practices for Multilingual Linked Open Data}
	\acro{BPSO}{Binary Particle-Swarm Optimization}
	\acro{BPMLOD}{Best Practices for Multilingual Linked Open Data}
	\acro{BFS}{Breadth-First-Search}
	\acro{BPE}{Byte Pair Encoding}
	\acro{BoW}{Bag-of-Words}
	\acro{CBD}{Concise Bounded Description}
	\acro{COG}{Content Oriented Guidelines}
	\acro{CSV}{Comma-Separated Values}
	\acro{CBMT}{Corpus-Based Machine Translation}
	\acro{CLIR}{Cross-Language Information Retrieval}
	\acro{DPSO}{Deterministic Particle-Swarm Optimization}
	\acro{DALY}{Disability Adjusted Life Year}

	\acro{ER}{Entity Resolution}
	\acro{EM}{Expectation Maximization}
	\acro{EBMT}{Example-Based Machine Translation}
	\acro{EBNF}{Extended Backus--Naur Form}
	\acro{EL}{Entity Linking}
	\acro{FAO}{Food and Agriculture Organization of the United Nations}
	\acro{GIS}{Geographic Information Systems}
	\acro{GHO}{Global Health Observatory}
	\acro{GRU}{Gated recurrent unit}
	\acro{HDI}{Human Development Index}
	\acro{ICT}{Information and communication technologies}
	\acro{IFRS}{International Financial Reporting Standards}
	\acro{ICD}{International Classification of Diseases}
	\acro{IT}{Information Technology}
    \acro{KB}{Knowledge Base}
    \acro{KG}{Knowledge Graph}
    \acro{KGE}{Knowledge Graph Embedding}
	\acro{LR}  {Language Resource}
	\acro{LD}  {Linked Data}
	\acro{LLOD}  {Linguistic Linked Open Data}
	\acro{LIMES}{LInk discovery framework for MEtric Spaces}
	\acro{LS}  {Link Specifications}
	\acro{LDIF}{Linked Data Integration Framework}
	\acro{LGD} {LinkedGeoData}
	\acro{LOD} {Linked Open Data}
	\acro{LSTM}{Long Short-Term Memories}
	\acro{MSE}{Mean Squared Error}
	\acro{MWE}{Multiword Expressions}
	\acro{MT}{Machine Translation}
	\acro{ML}{Machine Learning}
	\acro{MR}{Machine Reading}
	\acro{NIF}{Natural Language Processing Interchange Format}
	\acro{NIF4OGGD}{NLP Interchange Format for Open German Governmental Data}
	\acro{NLP}{Natural Language Processing}
	\acro{NER}{Named Entity Recognition}
	\acro{NMT}{Neural Machine Translation}
	\acro{NN}{Neural Network}
	\acro{NLG}{Natural Language Generation}
	\acro{NED}{Named Entity Disambiguation}
	\acro{NERD}{Named Entity Recognition and Disambiguation}
	\acro{NL}{Natural Language}
	\acro{NIF}{NLP Interchange Format}
	\acro{NIF4OGGD}{NLP Interchange Format for Open German Governmental Data}
	\acro{NLP}{Natural Language Processing}
	\acro{NER}{Named Entity Recognition}
	\acro{NEL}{Named Entity Linking}
	\acro{NE}{Named Entity}
	\acro{NN}{Neural Network}
	\acro{NLI}{Natural Language Inference}
	\acro{OSM}{OpenStreetMap}
	\acro{OWL}{Web Ontology Language}
	\acro{OOV}{out-of-vocabulary}
	\acro{PFM}{Pseudo-F-Measures}
	\acro{PSO}{Particle-Swarm Optimization}
	\acro{PBSMT}{Phrase-Based Statistical Machine Translation}
	
	\acro{QA}{Question Answering}
	\acro{RL}{Reinforcement Learning}
	\acro{RDF}{Resource Description Framework}
	\acro{RNN}{Recurrent Neural Network}
	\acro{ReLU}{rectified linear unit}
	
	\acro{SKOS}{Simple Knowledge Organization System}
	\acro{SWT}{Semantic Web Technologies}
	\acro{SW}{Semantic Web}
	\acro{SPARQL}{SPARQL Protocol and RDF Query Language}
	\acro{SRL}{Statistical Relational Learning}
	\acro{SVM}{Support Vector Machines}
    \acro{SGD}{Stochastic Gradient Descent}
    \acro{TBMT} {Transfer-Based Machine Translation}
	\acro{UML}{Unified Modeling Language}
	\acro{USL}{Ukrainian Sign Language}
	\acro{URI}{Uniform Resource Identifier}
	\acro{WHO}{World Health Organization}
	\acro{WKT}{Well-Known Text}
	\acro{W3C}{World Wide Web Consortium}
	\acro{WSD}{Word Sense Disambiguation}
    \acro{XML}{Extensible Markup Language}
	\acro{YPLL}{Years of Potential Life Lost}
\end{acronym}  
\newcommand{\RealNumbers}{\ensuremath{\mathbb{R}}\xspace}
\newcommand{\ComplexNumbers}{\ensuremath{\mathbb{C}}\xspace}
\newcommand{\Quaternions}{\ensuremath{\mathbb{H}}\xspace}
\newcommand{\Octonions}{\ensuremath{\mathbb{O}}\xspace}

\newcommand{\real}{\mathrm{Re}}

\newcommand{\ConvolutionWithProjection}{\text{conv}}

\DeclareMathOperator{\vect}{vec}

\newcommand{\approach}{\textsc{ConvQ}\xspace}
\newcommand{\baseapproach}{\textsc{QMult}\xspace}

\newcommand{\approachocto}{\textsc{ConvO}\xspace}
\newcommand{\baseapproachocto}{\textsc{OMult}\xspace}

\newcommand{\kg}{\ensuremath{\mathcal{G}}\xspace}
\newcommand{\emb}[1]{\ensuremath{\mathbf{e}_{#1}}}
\newcommand{\embdim}{\ensuremath{d}} % embedding dimensions for entities

\newcommand{\triple}[3]{(\texttt{#1}, \texttt{#2}, \texttt{#3})}
\newcommand{\pair}[2]{(\texttt{#1}, \texttt{#2})}

\newcommand{\entities}{\ensuremath{\mathcal{E}}\xspace}
\newcommand{\relations}{\ensuremath{\mathcal{R}}\xspace}

\newcommand{\scoreFunc}{\phi}

\title{Convolutional Hypercomplex Embeddings for Link Prediction}

\author[1]{\Name{Caglar Demir} \Email{caglar.demir@upb.de}
\\
\addr{Data Science Research Group, Paderborn University} \\
\Name{Diego Moussallem}\Email{diego.moussallem@upb.de}\\
\addr{Data Science Research Group, Paderborn University} \\
\addr{Globo, Rio de Janeiro, Brazil} \\
\Name{Stefan Heindorf}\Email{stefan.heindorf@upb.de}\\
\addr{Data Science Research Group, Paderborn University} \\
\Name{Axel-Cyrille {Ngonga Ngomo}}\Email{axel.ngonga@upb.de}\\
\addr{Data Science Research Group, Paderborn University} \\
}

\editors{Vineeth N Balasubramanian and Ivor Tsang}
\begin{document}

\maketitle
\begin{abstract}
Knowledge graph embedding research has mainly focused on the two smallest normed division algebras, $\mathbb{R}$ and $\mathbb{C}$. Recent results suggest that trilinear products of quaternion-valued embeddings can be a more effective means to tackle link prediction. In addition, models based on convolutions on real-valued embeddings often yield state-of-the-art results for link prediction. In this paper, we investigate a composition of convolution operations with hypercomplex multiplications. We propose the four approaches \baseapproach, \baseapproachocto, \approach and \approachocto  to tackle the link prediction problem. \baseapproach and \baseapproachocto can be considered as quaternion and octonion extensions of previous state-of-the-art approaches, including DistMult and ComplEx. \approach and \approachocto build upon \baseapproach and \baseapproachocto by including convolution operations in a way inspired by the residual learning framework. We evaluated our approaches on seven link prediction datasets including WN18RR, FB15K-237, and YAGO3-10. Experimental results suggest that the benefits of learning hypercomplex-valued vector representations become more apparent as the size and complexity of the knowledge graph grows. \approachocto outperforms state-of-the-art approaches on FB15K-237 in MRR, Hit@1 and Hit@3, while \baseapproach, \baseapproachocto, \approach and \approachocto outperform state-of-the-approaches on YAGO3-10 in all metrics. Results also suggest that link prediction performance can be further improved via prediction averaging. To foster reproducible research, we provide an open-source implementation of approaches, including training and evaluation scripts as well as pretrained models.%
\footnote{\url{https://github.com/dice-group/Convolutional-Hypercomplex-Embeddings-for-Link-Prediction}}
\end{abstract}
\begin{keywords}
Convolution, Hypercomplex Embeddings, Link Prediction, Residual Learning
\end{keywords}

\section{Introduction} % CD: Ready for the final submission
\label{sec: introduction}
% Context: Knowledge graph embeddings for link prediction.
Knowledge graphs represent structured collections of facts describing the world in the form of typed relationships between entities~\citep{hogan2020knowledge}. These collections of facts have been used in a wide range of applications, including web search, question answering, and recommender systems~\citep{nickel2015review}. However, most knowledge graphs on the web are far from complete. The task of identifying missing links in knowledge graphs is referred to as~\textit{link prediction}.~\ac{KGE} models have been particularly successful at tackling the link prediction task, among many others~\citep{nickel2015review}.

% Motivation for quaternions and octonions.
\ac{KGE} research has mainly focused on the two smallest normed division algebras---real numbers~($\RealNumbers$) and complex numbers~($\ComplexNumbers$)---neglecting the benefits of the larger normed division algebras---quaternions~($\Quaternions)$ and octonions~($\Octonions$). While~\cite{yang2015embedding} introduced the trilinear product of \emph{real}-valued embeddings $\langle \emb{h}, \emb{r}, \emb{t} \rangle$ of triples \triple{h}{r}{t} as a scoring function for link prediction,~\cite{trouillon2016complex} showed the usefulness of the Hermitian product of \emph{complex}-valued embeddings $\real(\langle \emb{h}, \emb{r}, \overline{ \emb{t}} \rangle)$. In contrast to the trilinear product of \emph{real}-valued embeddings, the Hermitian product is not symmetric and can be used to model antisymmetric relations since $\real(\langle \emb{h}, \emb{r}, \overline{ \emb{t}} \rangle) \ne \real(\langle \emb{t}, \emb{r}, \overline{ \emb{h}} \rangle)$. To further increase the expressivity,~\cite{zhang2019quaternion} proposed learning \emph{quaternion}-valued embeddings due to their benefits over \emph{complex}-valued embeddings.~\cite{zhang2021beyond} show that replacing a fully connected layer with a hypercomplex multiplication layer in a neural network leads to significant parameter efficiency without degenerating the predictive performance in many tasks.

% Motivation for convolutions
~\cite{dettmers2018convolutional},~\cite{balavzevic2019hypernetwork}, and ~\cite{demir2021convolutional} showed that convolutions are another effective means to increase the expressivity: the sparse connectivity property of the convolution operator endows models with parameter efficiency---unlike models simply increasing the embedding size which is not scalable to large knowledge graphs~\citep{dettmers2018convolutional}. Different configurations of the number of feature maps and the shape of kernels (1D or 2D) in the convolution operation are often explored to find the best ratio between expressiveness and parameter space size. 

% Our contribution
We investigate the use of convolutions 
on hypercomplex embeddings by proposing four models:
\baseapproach and \baseapproachocto can be considered hypercomplex extensions of DistMult~\citep{yang2015embedding} in~$\Quaternions$ and~$\Octonions$, respectively. In contrast to the state of the art~\citep{zhang2019quaternion}, we address the scaling effect of multiplication in $\Quaternions$ and $\Octonions$ by applying the batch normalization technique. Through the batch normalization technique, \baseapproach and \baseapproachocto are allowed to control the rate of normalization and benefit from its implicit regularization effect~\citep{ioffe2015batch}. Importantly,~\cite{lu2020dense} suggest that using solely unit quaternion-based rotations between head entity and relation limits the modeling capacity for various types of relations. \approach and \approachocto build upon \baseapproach and \baseapproachocto by including the convolution operator that is partially %in a way 
inspired by the residual learning framework~\citep{he2016deep}. \approach and \approachocto forge \baseapproach and \baseapproachocto with a 2D convolution operation and an affine transformation via the Hadamard product, respectively. By virtue of this architecture, we show that \approach can degenerate \baseapproach, ComplEx or DistMult, if such degeneration is necessary to further minimize the training loss (see~\Cref{eq:equivalance_qmult,eq:equivalance_complex}).
% Summary of experimental results
Experiments suggest that our models often achieve state-of-the-art performance on seven benchmark datasets (WN18, FB15K, WN18RR, FB15K-237, YAGO3-10, Kinship and UMLS). Superiority of our models against state-of-the-art models increases as the size and complexity of the knowledge graph grow. Our results also indicate that generalization performance of models can be further increased by applying ensemble learning.

\section{Related Work} % CD: Ready for the final submission
\label{sec:related-work}
In the last decade, a plethora of \ac{KGE} approaches have been successfully applied to tackle various tasks~\citep{nickel2015review}. In this section, we give a brief chronological overview of selected \ac{KGE} approaches. RESCAL computes a three-way factorization of a third-order adjacency tensor representing the input knowledge graph to compute scores for triples~\citep{nickel2011three}. RESCAL captures various types of relations in the input \ac{KG} but is limited in its scalability as it has quadratic complexity in the factorization rank~\citep{trouillon2017knowledge}. DistMult can be regarded as an efficient extension of  RESCAL with a diagonal matrix per relation to reduce the complexity of RESCAL~\citep{yang2015embedding}. DistMult performs poorly on antisymmetric relations, whereas it performs well on symmetric relations~\citep{trouillon2016complex}. ComplEx extends DistMult by learning representations in a complex vector space~\citep{trouillon2016complex}. ComplEx is able to infer both symmetric and antisymmetric relations via a Hermitian inner product of embeddings that involves the conjugate-transpose of one of the two input vectors.~\cite{lacroix2018canonical} designed two novel regularizers along with a data augmentation technique and propose ComplEx-N3 that can be seen as ComplEx with the N3 regularization. ConvE applies a 2D convolution operation to model the interactions between entities and relations~\citep{dettmers2018convolutional}. ConvKB extends ConvE by omitting the reshaping operation in the encoding of representations in the convolution operation~\citep{nguyen2017novel}. Similarly, HypER extends ConvE by applying relation-specific 1D convolutions as opposed to applying filters from concatenated subject and relation vectors~\citep{balavzevic2019hypernetwork}. TuckER employs the Tucker decomposition on the binary tensor representing the input knowledge graph triples~\citep{balavzevic2019tucker}. RotatE employs a rotational model taking predicates as rotations from subjects to objects in complex space via the element-wise Hadamard product~\citep{sun2019rotate}. By these means, RotatE performs well on composition relations where other approaches perform poorly. QuatE applies the quaternion multiplication followed by an inner product to compute scores of triples~\citep{zhang2019quaternion}.

\cite{nickel2017poincare},~\cite{balazevic2019multi}, and~\cite{demir2019physical} propose learning embeddings of input \ac{KG} such that distances between embeddings of entities and relations reflect their semantic similarity. 
\section{Link Prediction \& Hypercomplex Numbers}
% CD: Ready for the final submission
\label{sec:preliminaries}
\paragraph{Link Prediction.} 
Let \entities\ and \relations\ represent the sets of entities and relations. Then, a \ac{KG} can be formalised as a set of triples $\kg= \{\triple{h}{r}{t} \}  \subseteq \entities \times \relations \times \entities$ where each triple contains two entities $\texttt{h},\texttt{t} \in \entities$ and a relation $\texttt{r} \in \relations$. The link prediction problem is formalised by learning a scoring function $\scoreFunc:\entities \times \relations \times \entities \to \mathbb{R}$ ideally characterized by $\scoreFunc\triple{h}{r}{t} > \scoreFunc\triple{x}{y}{z}$ if $\triple{h}{r}{t}$ is true and $\triple{x}{y}{z}$ is not~\cite{dettmers2018convolutional,demir2021shallow}.
\paragraph{Hypercomplex Numbers.} 
The quaternions are a 4-dimensional normed division algebra~\citep{hamilton1844lxxviii}. A quaternion number $Q \in \Quaternions$ is defined as $ Q = a + b\textbf{i} + c\textbf{j}  + d\textbf{k}$ where $a, b, c , d $ are real numbers and $\textbf{i}, \textbf{j}, \textbf{k}$ are imaginary units satisfying Hamilton's rule: $ \textbf{i}^2 = \textbf{j}^2 = \textbf{k}^2 = \textbf{i}\textbf{j}\textbf{k} = -1$. The inner product of two quaternions is defined as
\begin{equation*}
\label{eq:quaternion_inner_product}
    Q_1 \cdot Q_2= \langle a_1 a_2 \rangle + \langle b_1 b_2 \rangle + \langle c_1 c_2\rangle + \langle d_1 d_2 \rangle.
\end{equation*}
The quaternion multiplication of $Q_1$ and $Q_2$ is defined as
\begin{equation*}
    \begin{split}
     Q_1 \otimes Q_2 &= (a_1 a_2 - b_1 b_2 -c_1 c_2 -d_1 d_2)\\ 
     & + (a_1 b_2 + b_1 a_2 + c_1 d_2 - d_1 c_2 ) \textbf{ i}\\
     & + (a_1 c_2 -b_1 d_2 + c_1 a_2 + d_1 b_2 ) \textbf{ j}\\
     & + (a_1 d_2 + b_1 c_2 - c_1 b_2 +d_1 a_2 ) \textbf{ k}.
    \end{split}
    \label{eq:quaternion_mul}
\end{equation*}
The quaternion multiplication is also known as the Hamilton product~\citep{zhang2021beyond}. 
For a $\embdim$-dimensional quaternion vector \mbox{a + b \textbf{i} + c \textbf{j} + d \textbf{k}} with $a,b,c,d \in \RealNumbers^\embdim$, the inner product and multiplication is defined accordingly. 
% $Q_1 \otimes Q_2$ has the effect of scaling $Q_1$ by the magnitude of $Q_2$ followed by a rotation~\citep{zhang2019quaternion}.
%
% A $\embdim$-dimensional quaternion-valued vector is defined as $\{a + b \textbf{i} + c \textbf{j} + d \textbf{k} : a,b,c,d \in \RealNumbers^ \embdim \}$. The inner product and quaternion multiplication of two quaternions similarly hold for quaternion-valued vectors.
% $Q^{\triangleleft}=Q/|Q|$ denotes a unit normalized quaternion with $|Q|=\sqrt{a^2 + b^2 + c^2 + d^2}$.  
The Octonions are an 8-dimensional algebra where an octonion number $O_1 \in \Octonions$ is defined as $O_1 = x_0 + x_1 \textbf{e}_1 + x_2 \textbf{e}_2 + \ldots + x_7 \textbf{e}_7$, where $\textbf{e}_1,\textbf{e}_2 \ldots \textbf{e}_7$ are imaginary units~\citep{baez2002octonions}. Their product ($\bigstar$), inner product ($\cdot$) and vector operations are defined analogously to quaternions.

The quaternion multiplication subsumes real-valued multiplication and enjoys a parameter saving with $1/4$ as compared to the real-valued matrix multiplication~\citep{parcollet2018quaternion,zhang2021beyond}. Leveraging such properties of quaternions in neural networks showed promising results in numerous tasks~\citep{zhang2021beyond,zhang2019quaternion,chen2020quaternion}. However,
%In turn, 
the octonion multiplication in neural networks and learning octonion-valued knowledge graph embeddings had not yet been fully explored.
\section{Convolutional Hypercomplex Embeddings}
\label{sec:approach}
% CD: Ready for the final submission
\paragraph{Motivation.}
\cite{dettmers2018convolutional} suggest that indegree and PageRank can be used to quantify the difficulty of predicting missing links in \acp{KG}. Results indicate that the superiority of ConvE becomes more apparent against DistMult and ComplEx as the complexity of the knowledge graph increases, i.e., indegree and PageRank of a \ac{KG} increase (see Table 6 in~\cite{dettmers2018convolutional}). In turn,~\cite{zhang2019quaternion} show that learning quaternion-valued embeddings via multiplicative interactions can be a more effective means of predicting missing links than learning real and complex-valued embeddings. Although learning quaternion-valued embeddings through multiplicative interactions yields promising results, the only way to further increase the expressiveness of such models is to increase the number of dimensions of embeddings. This does not scale to larger knowledge graphs~\citep{dettmers2018convolutional}. Increasing parameter efficiency while retaining effectiveness is a desired property in many applications~\citep{zhang2021beyond,trouillon2016complex,trouillon2017knowledge}.

Motivated by findings of aforementioned works, we investigate the composition of convolution operations with hypercomplex multiplications. The rationale behind this composition is to increase the expressiveness without increasing the number of parameters. This nontrivial endeavor is the keystone of embedding models~\citep{trouillon2016complex}. The sparse connectivity property of the convolution operation endows models with parameter efficiency,  which helps to scale to larger knowledge graphs. Additionally, different configurations of the number of kernels and their shapes can be explored to find the best ratio between expressiveness and the number of parameters. Although increasing the number of feature maps results in increasing the number of parameters, we are able to benefit from the parameter sharing property of convolutions~\citep{goodfellow2016deep}.
\paragraph{Approaches.}Inspired by the early works DistMult and ConvE, we dub our approaches \baseapproach, \baseapproachocto, \approach, and \approachocto, where ``Q'' represents the quaternion variant and ``O'' the octonion variant. Given a triple $\triple{h}{r}{t}$, $\baseapproach:\Quaternions \to \RealNumbers$ computes a triple score through the quaternion multiplication of head entity embeddings $\emb{h}$ and relation embeddings $\emb{r}$ followed by the inner product with tail entity embeddings $\emb{t}$ as 
\begin{equation}
\label{eq:qmult}
     \baseapproach(h,r,t) = \emb{h} \otimes \emb{r} \cdot \emb{t},
\end{equation}
where $\emb{h},\emb{r},\emb{t} \in \Quaternions^\embdim$. Similarly, $\baseapproachocto:\Octonions \to \RealNumbers$ performs
the octonion multiplication followed by the inner product as
\begin{equation}
\label{eq:omult}
     \baseapproachocto(h,r,t) = \emb{h} \bigstar \emb{r} \cdot \emb{t},
\end{equation}
where $\emb{h},\emb{r},\emb{t} \in \Octonions^\embdim$. Computing scores of triples in this setting can be illustrated in two consecutive steps: (1)~rotating \emb{h} through \emb{r} by applying quaternion/octonion multiplication,  and (2)~squishing ($\emb{h} \otimes \emb{r}$) and \emb{t} into a real number line by taking the inner product. During training, the degree between ($\emb{h} \otimes \emb{r}$) and \emb{t} is minimized provided $\triple{h}{r}{t} \in \kg$.

Motivated by the response of John T.\ Graves to W.\ R.\ Hamilton,\footnote{``If with your alchemy you can make three pounds of gold, why should you stop there?''~\cite{baez2002octonions}.} we combine 2D convolutions with \baseapproach and \baseapproachocto as defined
\begin{align}
  \label{eq:convq}
  \approach(h,r,t) &= \ConvolutionWithProjection(\emb{h},\emb{r})\circ (\emb{h} \otimes \emb{r} ) \cdot \emb{t},\\
  \label{eq:convo}
  \approachocto(h,r,t) &= \ConvolutionWithProjection(\emb{h},\emb{r})\circ (\emb{h} \bigstar \emb{r} ) \cdot \emb{t},
\end{align}
where $\ConvolutionWithProjection(\cdot,\cdot)
:\Quaternions^{2\embdim} \to \Quaternions^{\embdim}$ (respectively $:\Octonions^{2\embdim} \to \Octonions^{\embdim}$) is defined as 
\begin{equation}
    \ConvolutionWithProjection( \emb{h}, \emb{r} )  = f \big( \vect ( f ( [ \emb{h} , \emb{r}] \ast \omega ) ) \cdot \mathbf{W} +\textbf{b} \big).
\label{eq:convolution}
\end{equation}
$f(\cdot), \vect(\cdot), \ast, \omega$, $[\cdot,\cdot ]$, and ($\mathbf{W},\textbf{b}$) denote the rectified linear unit function, a flattening operation, convolution operation, kernel in the convolution, stacking operation,
%\footnote{in case of Octonions, this would yield real dimensions $4 \times d$ and in case of Sedenions $8 \times d$},
and an affine transformation, respectively.
\paragraph{Connection to ComplEx and DistMult.}
During training, $\ConvolutionWithProjection(\cdot,\cdot)$ can reduce its range into $\gamma \in 1$ if such reduction is necessary to further decrease the training loss. In the following \Cref{eq:equivalance_qmult,eq:equivalance_qmult_jk_zero,eq:equivalance_qmult_jk_open,eq:equivalance_qmult_jk_open_scaled,eq:equivalance_complex}, we elucidate the reduction of \approach into \baseapproach and ComplEx:
\begin{equation}
\approach\triple{h}{r}{t} = \gamma \circ (\emb{h} \otimes \emb{r} ) \cdot \emb{t}.
\label{eq:equivalance_qmult}
\end{equation}
\Cref{eq:equivalance_qmult} corresponds to \baseapproach provided that
$\ConvolutionWithProjection(\emb{h},\emb{r})=\gamma =1$. \approach can be further reduced into ComplEx by setting the imaginary parts \textbf{j} and \textbf{k} of \emb{h}, \emb{r} and \emb{t} to zero:
\begin{equation}
\gamma \circ \big( (a_h + b_h \textbf{i}) \otimes (a_r + b_r \textbf{i}) \big) \cdot (a_t + b_t \textbf{i}).
\label{eq:equivalance_qmult_jk_zero}
\end{equation}
Computing the quaternion multiplication of two quaternion-valued vectors corresponds to~\Cref{eq:equivalance_qmult_jk_open}:
\begin{equation}
\gamma \circ [(a_h \circ a_r - b_h\circ b_r) + (a_h\circ b_r + b_h\circ a_r) \textbf{i}] \cdot (a_t + b_t \textbf{i}).
\label{eq:equivalance_qmult_jk_open}
\end{equation}
The resulting quaternion-valued vector is scaled with $\gamma=[\gamma_1,\gamma_2]$:
\begin{equation}
[( \gamma_1 \circ a_h \circ a_r - \gamma_1 \circ b_h\circ b_r) +  ( \gamma_2 \circ a_h\circ b_r + \gamma_2 \circ b_h\circ a_r) \textbf{i}] \cdot (a_t + b_t \textbf{i}).
\label{eq:equivalance_qmult_jk_open_scaled}
\end{equation}
Through taking the inner product of the former vector with $(a_t + b_t \textbf{i})$, we obtain
\begin{equation}
\begin{split}
\approach\triple{h}{r}{t} =\langle \gamma_1, a_h, a_r, a_t\rangle \\
+\langle \gamma_2, b_h, a_r, b_t\rangle \\
+\langle  \gamma_2, a_h,b_r,b_t\rangle \\
-\langle \gamma_1, b_h, b_r, a_t\rangle,
\label{eq:equivalance_complex}
\end{split}
\end{equation}
where $\langle a, b, c, d\rangle = \sum_{k} a_k b_k c_k d_k$ corresponds to the multi-linear inner product. ~\Cref{eq:equivalance_complex} corresponds to ComplEx provided that $\gamma=1$. In the same way, \approach can be reduced into DistMult by setting all imaginary parts \textbf{i}, \textbf{j}, \textbf{k} to zero for
\emb{h}, \emb{r}, and \emb{t} % $e_h$, $e_r$ and $e_t$
yielding
\begin{equation}
\approach\triple{h}{r}{t} =\langle \gamma_1, a_h, a_r, a_t\rangle .
\end{equation}
\paragraph{Connection to residual learning.}
The residual learning framework facilitates the training of deep neural networks. A simple residual learning block consists of two weight layers denoted by $\mathcal{F}(x)$ and an identity mapping of the input $x$ (see Figure 2 in~\cite{he2016deep}). Increasing the depth of a neural model via stacking residual learning blocks led to significant improvements in many domains. In our setting, $\mathcal{F}(\cdot)$ and $x$ correspond to $\ConvolutionWithProjection( \cdot, \cdot )$ and $[\emb{h}, \emb{r}]$, respectively. We replaced the identity mapping of the input with the hypercomplex multiplication. To scale the output, we replaced the elementwise vector addition with the Hadamard product. By virtue of such inclusion, 
\approach and \approachocto are endowed with the ability of controlling the impact of $\ConvolutionWithProjection( \cdot, \cdot )$ on predicted scores as shown in \Cref{eq:equivalance_complex}. Ergo, 
the gradients of loss (see \Cref{eq:loss}) w.r.t. head entity and relation embeddings can be propagated in two ways, namely, via $\ConvolutionWithProjection(\emb{h},\emb{r})$ or hypercomplex multiplication. Moreover, the number of feature maps and the shape of kernels can be used to find the best ratio between expressiveness and the number of parameters. Hence, the expressiveness of models can be adjusted without necessarily increasing the embedding size. Although increasing the number of feature maps results in increasing the number of parameters in the model, we are able to benefit from the parameter sharing property of convolutions.
\section{Experimental Setup}
\label{sec:experiments}
\subsection{Datasets}
We used seven datasets: WN18RR, FB15K-237, YAGO3-10, FB15K, WN18, UMLS and Kinship. An overview of the datasets is provided in~\Cref{all_dataset_info}. The latter four datasets are included for the sake of the completeness of our evaluation. \cite{dettmers2018convolutional} 
suggest that indegree and PageRank can be used to indicate difficulty of performing link prediction on an input~\ac{KG}. In our experiments, we are particularly interested in link prediction results on complex \acp{KG}.
As commonly done, we augment the datasets by adding reciprocal triples \triple{t}{r$^{-1}$}{h}~\citep{dettmers2018convolutional,balavzevic2019hypernetwork,balavzevic2019tucker}. For link prediction based on only tail entity ranking experiments (see \Cref{table:link-prediction-tail}), we omit the data augmentation on the test set, as in~\citep{bansal2019a2n}.%as similarly done by~\cite{bansal2019a2n}. 
\begin{table}[tbp]
    \caption{Overview of datasets in terms of entities, relations, average node degree plus/minus standard deviation.}
    \centering
    \footnotesize
    \setlength{\tabcolsep}{6pt}
    %\scalebox{0.90}{%
    \begin{tabular}{@{}l r r r@{} r r r@{}}
    \toprule
    \bfseries Dataset & \multicolumn{1}{c}{$|\entities|$}&  $|\relations|$ & \multicolumn{1}{c}{Degree} & $|\kg^{\text{Train}}|$ & $|\kg^{\text{Val.}}|$ &  $|\kg^{\text{Test}}|$\\
    \midrule
    YAGO3-10  & 123,182 &     37 & \phantom{0}9.6$\pm$8.7\phantom{0}  & 1,079,040 &  5,000 &  5,000 \\
    FB15K     &  14,951 &  1,345 &           32.5$\pm$69.5            &   483,142 & 50,000 & 59,071 \\
    WN18      &  40,943 &     18 &            3.5$\pm$7.7\phantom{0}  &   141,442 &  5,000 &  5,000 \\
    FB15k-237 &  14,541 &    237 &           19.7$\pm$30\phantom{.0}  &   272,115 & 17,535 & 20,466 \\
    WN18RR    &  40,943 &     11 & \phantom{0}2.2$\pm$3.6\phantom{0}  &    86,835 &  3,034 &  3,134 \\
    KINSHIP   &     104 &    25  &           82.2$\pm$3.5\phantom{0}  &     8,544 &  1,068 &  1,074 \\
    UMLS      &     135 &    46  &           38.6$\pm$32.5            &     5,216 &    652 &    661 \\
    \bottomrule
    \end{tabular}
    \label{all_dataset_info}%}
\end{table}
\subsection{Training and optimization}
We apply the standard training strategy as \cite{dettmers2018convolutional,balavzevic2019hypernetwork,balavzevic2019tucker}:
Following the KvsAll training procedure,%
\footnote{Here, we follow the terminology of~\cite{ruffinelli2019you}.}
for a given pair \pair{h}{r}, we compute scores for all $x \in \entities$ with $\scoreFunc(h,r,x)$ and apply the logistic sigmoid function $\sigma(\scoreFunc(h,r,x))$. Models are trained to minimize the binary cross entropy loss function, where $\hat{\mathbf{y}} \in \mathbb{R}^{|\entities|}$ and $\mathbf{y} \in [0,1]^{|\entities|}$ denote
the predicted scores and binary label vector, respectively. 
\begin{equation}
L = -\frac{1}{|\entities|}\sum\limits_{i=1}^{|\entities|} (\mathbf{y}^{(i)} \text{log}(\hat{\mathbf{y}}^{(i)}) + (1-\mathbf{y}^{(i)}) \text{log}(1-\hat{\mathbf{y}}^{(i)})).
\label{eq:loss}
\end{equation}
We employ the Adam optimizer~\citep{kingma2014adam}, dropout~\citep{srivastava2014dropout}, label smoothing and batch normalization~\citep{ioffe2015batch}, as in the literature~\citep{balavzevic2019hypernetwork,balavzevic2019tucker,dettmers2018convolutional,demir2021shallow}. Moreover, we selected hyperparameters of our approaches by random search based on validation set performances~\citep{balavzevic2019tucker}.
Notably, we did not search a good random seed for the random number generator, and we fixed the seed to 1 throughout our experiments.
\subsection{Evaluation}
We employ the standard metrics \textit{filtered} \ac{MRR} and hits at N (H@N)  for link prediction~\citep{dettmers2018convolutional,balavzevic2019hypernetwork}. For each test triple $\triple{h}{r}{t}$, we construct its reciprocal $\triple{t}{r$^{-1}$}{h}$ and add it into $\kg^{\text{test}}$, which is a common technique to decrease the computational cost during testing~\citep{dettmers2018convolutional}. Then, for each test triple $\triple{h}{r}{t}$, we compute the score of $\triple{h}{r}{x}$ triples for all $x \in \entities$ and calculate the filtered ranking $rank_t$ of the triple having $t$. Then we compute the \ac{MRR}: $\frac{1}{|\kg^{\text{test}}|} \sum_{\triple{h}{r}{t} \in \kg^{\text{test}}} \frac{1}{rank_t}$. Consequently, given a $\triple{h}{r}{t} \in \kg^{\text{test}}$, we compute ranks of missing entities based on \emph{the rank of head and tail entities} analogously to
~\citep{dettmers2018convolutional,balavzevic2019tucker,balavzevic2019hypernetwork}. For the sake of completeness, we also report link prediction performances based on \emph{only tail rankings}, i.e., without including triples with reciprocal relations into test data, as in~\cite{bansal2019a2n}.
\subsection{Implementation Details and Reproducibility}
We implemented and evaluated our approach in the framework provided by~\cite{balavzevic2019tucker}. To alleviate the hardware requirements for the reproducibility, we provide hyperparameter optimization, training and evaluation scripts along with pretrained models at the project page. Experiments were conducted on a single NVIDIA GeForce RTX 3090.
\section{Results}
\label{sec:results}

\Cref{table:link-prediction-main} reports link prediction results on the WN18RR, FB15K-237, and YAGO3-10 datasets. Overall, the superior performance of our approaches becomes more and more apparent as the size and complexity of the knowledge graphs grow. On the smallest benchmark dataset (WN18RR), \baseapproach, \baseapproachocto, \approach and \approachocto outperform many approaches including DistMult, ConvE and ComplEx in all metrics. However, QuatE, TuckER, and RotatE yield the best performance. On the second-largest benchmark dataset (FB15K-237 is $\textbf{3.1}\times$ larger than WN18RR), \approachocto outperforms all state-of-the-art approaches in 3 out of 4 metrics. Additionally, \baseapproach and \approach outperform all state-of-the-art approaches except for TucKER in terms of MRR, H@1 and H@3. On the largest benchmark dataset (YAGO3-10 is $\textbf{12.4}\times$ larger than WN18RR), 
\baseapproach, \approachocto, \approach outperform all approaches in all metrics. Surprisingly, \baseapproach and \baseapproachocto reach the best and second-best performance in all metrics, whereas \approachocto does not perform particularly well compared to our other approaches. \approachocto outperforms \baseapproach, \baseapproachocto, and \approach
in 8 out of 12 metrics, whereas
\baseapproach yields better performance on YAGO3-10. Overall, these results suggest that
superiority of learning hypercomplex embeddings becomes more apparent as the size and complexity of the input knowledge graph increases, as measured by indegree (see \Cref{all_dataset_info}) and PageRank (see Table 6 in~\cite{dettmers2018convolutional}). In~\Cref{table:link-prediction-main-param}, we compare some of the best performing approaches on WN18RR, FB15K-237 and YAGO3-10 in terms of the number of trainable parameters. Results indicate that our approaches yield competitive (if not better) performance on all benchmark datasets. 
\begin{table}[tbp]
    \caption{Link prediction results on WN18RR, F15K-237 and YAGO3-10. Results are obtained from corresponding papers. Bold and underlined entries denote best and second-best results. The dash (-) denotes values missing in the papers.}
    \label{table:link-prediction-main}
    \centering
    \footnotesize
    \setlength{\tabcolsep}{2.3pt}
\begin{tabular}{@{}l @{\hskip 0pt} c c c c c @{\hskip 8pt} c c c c c @{\hskip 8pt} c c c c c}
  \toprule
  &\multicolumn{4}{c}{\textbf{WN18RR}} & \multicolumn{4}{c}{\textbf{FB15K-237}} & \multicolumn{4}{c}{\textbf{YAGO3-10}} \\
  \cmidrule(lr){2-5} \cmidrule(lr){6-9} \cmidrule(l){10-13}
  & MRR  & @1  & @3 & @10  & MRR  & @1 & @3 & @10 & MRR  & @1 & @3 & @10  \\
  \midrule
  TransE~\citep{ruffinelli2019you}            & .228 &  .053  &  .368  & .520    & .313 & .221 & .347 & .497 &- &- &- &-\\
  ConvE~\citep{ruffinelli2019you}             & .442 &  .411  &  .451  &  .504    & .339 & .248 & .359 & .521 &- &- &- &-\\
  TuckER~\citep{balavzevic2019tucker}         & .470 & \textbf{.443} & .482 & .526    & \underline{.358}    & \underline{.266}    & \underline{.394}    & \textbf{.544} &- &- &- &-\\
  A2N~\citep{bansal2019a2n}                   & .450 & .420 & .460 & .510 & .317 & .232 & .348 & .486 & - & - & - &-\\
  QuatE~\citep{zhang2019quaternion}           & \textbf{.482}  & .436 & \textbf{.499} & \textbf{.572} &.311 & .221 & .342 & .495 & - &- &- &-\\
  HypER~\citep{balavzevic2019hypernetwork}    & .465 & .436 & .477 & .522 &.341    &.252    & .376    & .520   &  .533 & .455 & .580 &  .678\\
  DistMult~\citep{dettmers2018convolutional}  & .430 & .390 & .440 & .490 & .240 & .160 & .260 & .420 & .340 & .240 & .380 & .540\\
  ConvE~\citep{dettmers2018convolutional}     & .430 & .400 & .440 & .520 & .335 & .237 & .356 & .501  &  .440 & .350 & .490 &  .620\\
  ComplEx~\citep{dettmers2018convolutional}   & .440 & .410 & .460 & .510 & .247 & .158 & .275 & .428 & .360 &   .260 & .400 & .550\\
  REFE~\citep{chami2020low}                   & .455 & .419 & .470 & .521 & .302 & .216 & .330 & .474 &  .370 & .289 & .403 &  .527\\
  ROTE~\citep{chami2020low}                   & .463 & .426 & .477 & .529 & .307 & .220 & .337 & .482 &  .381 & .295 & .417 &  .548\\
  ATTE~\citep{chami2020low}                   & .456 & .419 & .471 & .526 & .311 & .223 & .339 & .488  &  .374 & .290 & .410 &  .538\\
  ComplEx-N3~\citep{chami2020low}             & .420  & .390 & .420 & .460 & .294 & .211 & .322 & .463 &  .336 & .259 & .367 &  .484\\
  MuRE~\citep{chami2020low}                   & .458  & .421 & .471 & .525 & .313 & .226 & .340 & .489 &  .283 & .187 & .317 &  .478\\
  RotatE~\citep{sun2019rotate}                 & \underline{.476}  & \underline{.428} & \underline{.492} & \underline{.571} & .338 & .241 & .375 & .533 &  .495 & .402 & .550 &  .670\\
  \midrule
  \baseapproach                                & .438 & .393  & .449 & .537 & .346 & .252 & .383 & .535 & \textbf{.555} & \textbf{.475} & \textbf{.602} & \textbf{.698}\\
  \baseapproachocto                            & .449  & .406 & .467 & .539 & .347 & .253 & .383 & .534 & \underline{.543} & \underline{.461} & \underline{.592}& \underline{.692}\\
  \approach                                    & .457 & .424  & .470 & .525 & .343 & .251 & .376 & .528 & .539& .459& .587&.687\\
  \approachocto                                & .458 & .427  & .473 & .521 & \textbf{.366} & \textbf{.271} & \textbf{.403} & \underline{.543} & .489& .395& .546&.664\\
\bottomrule
\end{tabular}%}
\end{table}

\begin{table}
\caption{Number of parameter comparisons on the WN18RR, FB15K-237 and YAGO3-10 datasets. The dash (-) denotes values missing in the papers.}
\label{table:link-prediction-main-param}
\centering
\footnotesize
%\scalebox{0.90}{%
\begin{tabular}{@{}l @{\hskip 0pt} c c c c c @{\hskip 8pt} c c c c c @{\hskip 8pt} c c c c c}
  \toprule
  &\multicolumn{1}{c}{\textbf{WN18RR}} & \multicolumn{1}{c}{\textbf{FB15K-237}} & \multicolumn{1}{c}{\textbf{YAGO3-10}} \\
  \midrule
  QuatE~\citep{zhang2019quaternion}           & 16.38M & 5.82M &-\\
  RotatE~\citep{sun2019rotate}                & 40.95M & 29.32M & 123.22M\\
  \midrule
  \baseapproach                               & 16.38M & 6.01M & 49.30M \\
  \baseapproachocto                           & 16.38M & 6.01M & 49.30M \\
  \approach                                   & 21.51M & 11.13M & 54.42M\\
  \approachocto                               & 21.51M & 11.13M & 54.42M\\
\bottomrule
\end{tabular}%}
\end{table}
\subsection{Ensemble Learning} 
\begin{table}[tbp]
\caption{Link prediction results via ensembling models.}
\label{table:main_results_ensemble}
% Results are double-checked.
\centering
\footnotesize
%\scalebox{0.90}{%
\setlength{\tabcolsep}{5pt}
\begin{tabular}{@{}l c c c c c c c c c c c c@{}}
  \toprule
  &\multicolumn{4}{c}{\textbf{WN18RR}} & \multicolumn{4}{c}{\textbf{FB15K-237}} & \multicolumn{4}{c}{\textbf{YAGO3-10}}\\
  \cmidrule(l){2-5} \cmidrule(l){6-9} \cmidrule(l){10-13}
                                    & MRR           & @1    & @3   & @10   & MRR         & @1 & @3 & @10 & MRR  & @1 & @3 & @10\\
  \midrule
    \textsc{Q-OMult}                & .444          & .399  & .458 & .544  & .356        & .260 & .393 & .545 & .557 & \underline{.478} & .601 &.700\\
    \baseapproach-\approach         & .446          & .406  & .455 & \underline{.538}  & .357        & .263 & .392 & .546 & \textbf{.561} & \textbf{.483} & \textbf{.606} & \textbf{.703}\\
    \baseapproach-\approachocto     & .449          & \underline{.410}  & \underline{.459} & .536  &\textbf{.372}& \underline{.275} & \textbf{.411} & \underline{.564}& .543 & .460 & .594 & .693\\
    \baseapproachocto-\approach     & .444          & .403  & .453 & .537  & .357        & .262 & .391 & .547 & \underline{.558} & \underline{.478} & \underline{.602} & .700\\
    \baseapproachocto-\approachocto & \underline{.462}          & \textbf{.425}  & \textbf{.475} & \textbf{.539}  &\textbf{.372}&\textbf{.277}&\textbf{.411} & \underline{.564}& .535 & .450 & .588 & .692\\
    \textsc{ConvQ-O-OMult}          & \textbf{.463} & \textbf{.425}  & \textbf{.475} & \textbf{.539}  &\textbf{.372}& \underline{.275}& \textbf{.411}  & \textbf{.567} & .552 & .470 &.599 &\underline{.702}\\
\bottomrule
\end{tabular}%}
\end{table}
~\Cref{table:main_results_ensemble} reports link prediction results of ensembled models on benchmark datasets. Averaging the predicted scores of models improved the performance by circa 1--2\% in \ac{MRR}. These results suggest that performance may be further improved through optimizing the impact of each model in the ensemble.
\subsection{Impact of Tail Entity Rankings}
During our experiments, we observed that models often perform more accurately in predicting missing tail entities compared to predicting missing head entities, which was also observed in~\cite{bansal2019a2n}.~\Cref{table:link-prediction-tail} indicates that \ac{MRR} performance based on \emph{only tail entity rankings} are on average absolute $10\%$ higher than
\ac{MRR} results based on \emph{head and tail entity rankings} on FB15K-237 while such difference was not observed on WN18RR.
\begin{table}[tbp]
\caption{Link prediction results based on only tail entity rankings.}
\label{table:link-prediction-tail}
\centering
\footnotesize
\setlength{\tabcolsep}{5.4pt}
%\scalebox{.90}{
\begin{tabular}{@{}l c c c c c c c c c c c c@{}}
\toprule
  &\multicolumn{4}{c}{\textbf{WN18RR}} & \multicolumn{4}{c}{\textbf{FB15K-237}} & \multicolumn{4}{c}{\textbf{YAGO3-10}}\\
   \cmidrule(l){2-5} \cmidrule(l){6-9} \cmidrule(l){10-13}
                                                       & MRR            & @1            & @3          & @10       & MRR  & @1 & @3 & @10 & MRR  & @1 & @3 & @10\\
  \midrule
    DistMult                                            &.430            &.410            &.440          &.480    & .370 & .275 & .417 & .568 & - & - & - & -\\
    ComplEx                                             &.420            &.380            &.430          &.480    & .394 & .303 & .434 & .572 & - & - & - & -\\
    ConvE                                               &.440            &.400            &.450          &.520    & .410 & .313 & .457 & .600 & - & - & - & -\\
    MINERVA                                             &.450            &.410            &.460          &.510    & .293 & .217 & .329 & .456 & - & - & - & -\\
    A2N                                                 &\textbf{.490}   &\textbf{.450}   &\textbf{.500}  &.550    & .422 & .328 & .464 & .608 & - & - & - & -\\
  \midrule
  \baseapproach                                         &.451            &.403            &.472          & .553 &.439 &.341&.485&.636& .692 & .626 & \underline{.736} & .799\\
  \baseapproachocto                                     &.461            &.414            &.482          &\underline{.559} & .440 & .340 &.486&.636& .689 & .623 & .733 & .801\\
  \approach                                             &.470            &.437            &.482          &.538 & .441 & .344 & .482 & .632& .674 & .612 & .715 & .783\\
  \approachocto                                         &.473            &\underline{.442}            &.491          &.535 & .465 & \underline{.367} & .512 & .654 & .622 & .535 & .682 & 774\\
  \midrule
  \textsc{Q-OMult}                  & .457 & .401  & .48 & \underline{.559} &  .448 & .347 & .495 & .644      & \underline{.696} & \underline{.632} & .734 & \textbf{.804}\\
  \baseapproach-\approach         & .466 & .424  & .480 & .557 &  .451 & .353 & .496 & .646      & \textbf{.697} & \textbf{.635} & \textbf{.737} & \underline{.803}\\
  \baseapproach-\approachocto     & .474 & .435  & .487 & .558 &  .467 & \underline{.367} & \underline{.516} & \textbf{.662}      & .680 & .610 & .727 & .800\\
  \baseapproachocto-\approach     & .471 & .430  & .487 & \textbf{.560} &  .452 & .354 & .495 & .647      & \underline{.696} & \underline{.632} & .735 & \underline{.803}\\
  \baseapproachocto-\approachocto & .476 & .436  & .488 & \underline{.559} &  \underline{.466} & .366 &.515& \textbf{.662}      & .676 & .602 & .724 & \underline{.803}\\
    \textsc{ConvQ-O}  & \underline{.477} & \underline{.442}  & \underline{.494} & .548 &  \textbf{.468} & \textbf{.370} & \textbf{.515} & \underline{.661}      & .675 & .603 & .724 & .795\\
\bottomrule
\end{tabular}%}
\end{table}

\begin{table}[tbp]
\caption{MRR link prediction results per relations on WN18RR. Ensemble refers to averaging predictions of  \approach-\approachocto-\baseapproachocto.}
\label{table:mrr_per_rel_wn18rr_comparision}
\centering
\footnotesize
\setlength{\tabcolsep}{4.5pt}
%\scalebox{.90}{%
\begin{tabular}{@{}l@{\hskip -6pt}ccccccc@{}}
\toprule
\bfseries Relation Name & \bfseries Rel.\ Type & \bfseries  RotatE &\bfseries  \baseapproach & \bfseries \approach & \bfseries \baseapproachocto & \bfseries \approachocto & \bfseries Ensemble \\ \midrule
hypernym                      & S &\textbf{.15}  & .10            &\underline{.14}         &.11              &.13               & .13 \\
instance\_hypernym            & S &.32           &.35            &\underline{.37}          &.36            &\underline{.37}               & \textbf{.39} \\
member\_meronym               & C&\textbf{.23}  &\underline{.22}            &.20          &\textbf{.23}            &.20               & \textbf{.23} \\
synset\_domain\_topic\_of     & C&\textbf{.34}  &.31            &.31          &.32            &\underline{.33}   & \textbf{.34} \\
has\_part                     & C&\underline{.18}           &\textbf{.19}            &.17          &\underline{.18}            &\underline{.18}               & \textbf{.19} \\
member\_of\_domain\_usage     & C&\underline{.32} & .29            &.28           &.27        &\textbf{.33}               & .29 \\
member\_of\_domain\_region    & C&.20          &.25            &\textbf{.38}          &.30    &\underline{.37}               & \textbf{.38}          \\
derivationally\_related\_form & R&.95           &\textbf{.98}  &\textbf{.98} &\textbf{.98}    &\textbf{.98}      & \textbf{.98} \\
also\_see                     & R &.59           &\textbf{.67}            &.65          &\underline{.66}            &\underline{.66}               & \underline{.66} \\
verb\_group                   & R&\underline{.94}          &\textbf{1.0} &\textbf{1.0}  &\textbf{1.0}    &\textbf{1.0}      & \textbf{1.0}          \\
similar\_to                   & R&\textbf{1.0} &\textbf{1.0}  &\textbf{1.0}  &\textbf{1.0}    &\textbf{1.0}      & \textbf{1.0} \\ \bottomrule
\end{tabular}%}
\end{table}
\begin{table}[tbp]
\caption{Link prediction results depending on the direction of prediction (head vs. tail prediction) on WN18RR. Ensemble refers to averaging predictions of \approach-\approachocto-\baseapproachocto. }
\label{table:per_rel_wn18rr_direction}
\centering
\setlength{\tabcolsep}{6pt}
\footnotesize
\scalebox{0.80}{%
\begin{tabular}{@{}lccccc@{}}
\toprule
& \textbf{\baseapproach} & \textbf{\approach} & \textbf{\baseapproachocto} & \textbf{\approachocto} & \textbf{Ensemble} \\
\midrule
\triple{h}{r}{x} \\ \midrule
hypernym                               & .12             & \textbf{.18}      & .13               & \textbf{.18}             & \underline{.17}\\
instance\_hypernym                     & .53             & .56      & .53               & \underline{.57}             & \textbf{.58} \\
member\_meronym                        & \textbf{.17}             & .09      & \underline{.16}               & .08             & .14\\
synset\_domain\_topic\_of              & .49             & .47      & \underline{.51}               & \textbf{.52}             & \textbf{.52} \\
has\_part                              & \textbf{.15}             & .12      & .15               & .14             & .14 \\
member\_of\_domain\_usage              & .04             & .02      & \underline{.07}               & \textbf{.08}             & .05 \\
member\_of\_domain\_region             & \underline{.05}             & \textbf{.06}      & \underline{.05}               & \textbf{.06}             & \underline{.05} \\
derivationally\_related\_form          & \underline{.98}             & \underline{.98}      & \underline{.98}               & \underline{.98}             & \underline{.98} \\
also\_see                              & \textbf{.67}             & .63      & \underline{.65}               & .63             & .63 \\
verb\_group                            & \textbf{1.0}             & \textbf{1.0}      & \textbf{1.0}               & \textbf{1.0}             &  \textbf{1.0} \\
similar\_to                            & \textbf{1.0}             & \textbf{1.0}      & \textbf{1.0}               & \textbf{1.0}             & \textbf{1.0} \\
\midrule
\triple{x}{r}{t}\\ \midrule
hypernym                      & .07             & \underline{.09}      & \underline{.09}               & .08             & \textbf{.10}\\
instance\_hypernym            & .17             & \underline{.18}      & \textbf{.19}               & .17             & \textbf{.19} \\
member\_meronym               & .27             & .31      & .29               & \textbf{.33}             & \underline{.32} \\
synset\_domain\_topic\_of     & .13             & \underline{.14}      & .13               & \textbf{.15}             & \textbf{.15}\\
has\_part                     & \underline{.22}             & \underline{.22}      & \underline{.22}               & \underline{.22}             & \textbf{.24} \\
member\_of\_domain\_usage     & .53             & \underline{.54}      & .47               & \textbf{.59}             & \underline{.54}\\
member\_of\_domain\_region    & .45             & \underline{.69}      & .55               & .67             & \textbf{.70} \\

derivationally\_related\_form & \underline{.98}             & \underline{.98}      & \underline{.98}               & \textbf{.99}             & \underline{.98} \\
also\_see                     & \underline{.68}             & .67      & .66               & \textbf{.69}             & \underline{.68} \\
verb\_group                   & \textbf{1.0}    & \textbf{1.0}      & \textbf{1.0}               & \textbf{1.0}             & \textbf{1.0}\\
similar\_to                   &\textbf{1.0}    &\textbf{1.0}      &\textbf{1.0} & \textbf{1.0}             & \textbf{1.0}\\
\bottomrule
\end{tabular}}
\end{table}
\subsection{Link Prediction Per Relation and Direction}
We reevaluate link prediction performance of some of the best-performing models from~\Cref{table:link-prediction-main} in~\Cref{table:mrr_per_rel_wn18rr_comparision,table:per_rel_wn18rr_direction}.~\cite{allen2021interpreting} distinguish three types of relations: Type~S relations are specialization relations such as \texttt{hypernym}, type~C denote so-called generalized context-shifts and include \texttt{has\_part} relations, and type R relations include so-called highly-related relations such as \texttt{similar\_to}. Our results show that our approaches accurately rank missing tail and head entities for type R relations. For instance, our
approaches perfectly rank ($1.0$ \ac{MRR}) missing entities of symmetric relations (\texttt{verb\_group} and \texttt{similar\_to}). However, the direction of entity prediction has a significant impact on the results for non-symmetric type C relations. For instance, \ac{MRR} performance of \baseapproach, \approach, \baseapproachocto and \approachocto vary by up to absolute 0.63 for the relation \texttt{member\_of\_domain\_region}. The low performance of \texttt{hypernym} (type S) may stem from the fact that there are 184 triples in the test split of WN18RR where \texttt{hypernym} occurs with entities of which at least one did not occur in the training split~\citep{demir2021out}. Models often perform poorly on type~C relations but considerably better on type R relations corroborating findings by~\cite{allen2021interpreting}.
\subsection{Batch vs.\ Unit Normalization}
We investigate the effect of using batch-normalization, instead of unit normalization as previously proposed by~\cite{zhang2019quaternion}. ~\Cref{table:umls_kinshp} indicates that the scaling effect of hypercomplex multiplications can be effectively alleviated by using the batch normalization technique. Replacing unit normalization with the batch normalization technique allows benefiting (1)~from its regularization effect and (2)~from its numerical stability. % In our experiments, we observed that applying the dropout technique after unit normalization lead to numeric instability. 
Through batch normalization, our models are able to control the rate of normalization and benefit from its implicit regularization effect~\citep{ioffe2015batch}.
\begin{table}[tbp]
	\centering
	\caption{Batch normalization vs.\ unit normalization for link prediction.}
	\scalebox{0.90}{%
	\setlength{\tabcolsep}{6pt}
	\footnotesize
    \begin{tabular}{@{}lcccc@{\hskip 10pt}cccc@{}}
    \toprule 
    &\multicolumn{4}{c}{\textbf{Kinship}}&\multicolumn{4}{c}{
    \textbf{UMLS}}\\ 
    \cmidrule(lr{10pt}){2-5} \cmidrule{6-9}
                                                        & MRR & H@1 & @3 & @10 & MRR & @1 & @3 & @10\\
    \midrule
	GNTP-Standard~\cite{minervini2020differentiable}   & .72 & .59 & .82 & .96  & .80 & .70 & .88 & .95\\
	GNTP-Attention~\cite{minervini2020differentiable}  & .76 & .64 & .85 & .96  & .86 & .76 & .95 & .98\\
	NTP~\cite{minervini2020differentiable}             & .35 & .24 & .37 & .57  & .80 & .70 & .88 & .95\\
	NeuralLP~\cite{minervini2020differentiable}        & .62 & .48 & .71 & .91  & .78 & .64 & .87 & .96\\
	MINERVA~\cite{minervini2020differentiable}         & .72 & .60 & .81 & .92  & .83 & .73 & .90 & .97\\
	ConvE~\cite{dettmers2018convolutional}             & .83 & .74 & .92 & \underline{.98}  & .94 & \underline{.92} & \underline{.96} & \underline{.99}\\
	\midrule
	\baseapproach (batch)                                           & \textbf{.88} & \textbf{.81} & \textbf{.94} & \textbf{.99}  & \textbf{.96} & \textbf{.93} & \textbf{.98} & \textbf{1.0}\\
	\baseapproach (unit)                              & .69 & .58 & .78 & .90  & .77 & .69 & .82 & .93\\ \midrule
	\baseapproachocto (batch)                                         & \underline{.87} & \underline{.80} & \textbf{.94} & \textbf{.99}  & \underline{.95} & .91 & \textbf{.98} & \textbf{1.0}\\
	\baseapproachocto (unit)                         & .69 & .57 & .77 & .89  & .76 & .66 & .82 & .93\\ \midrule
	\approach (batch)                                                & .86 & .77 & \underline{.93} & \underline{.98}  & .92 & .86 & \textbf{.98} & \textbf{1.0}\\
	\approach (unit)                                 & .61 & .49 & .68 & .85  & .55 & .45 & .59 & .75\\ \midrule
	\approachocto  (batch)                             & .86 & .77 & \underline{.93} & \underline{.98}  & .90 & .82 & \textbf{.98} & \textbf{1.0}\\
	\approachocto (unit)                             & .65 & .53 & .72 & .86  & .56 & .46 & .61 & .78\\
    \bottomrule
\end{tabular}}
 \label{table:umls_kinshp}
\end{table}

\subsection{Convergence on YAGO3-10}
\begin{figure}[tb]
    \centering
    \includegraphics[scale=0.6]{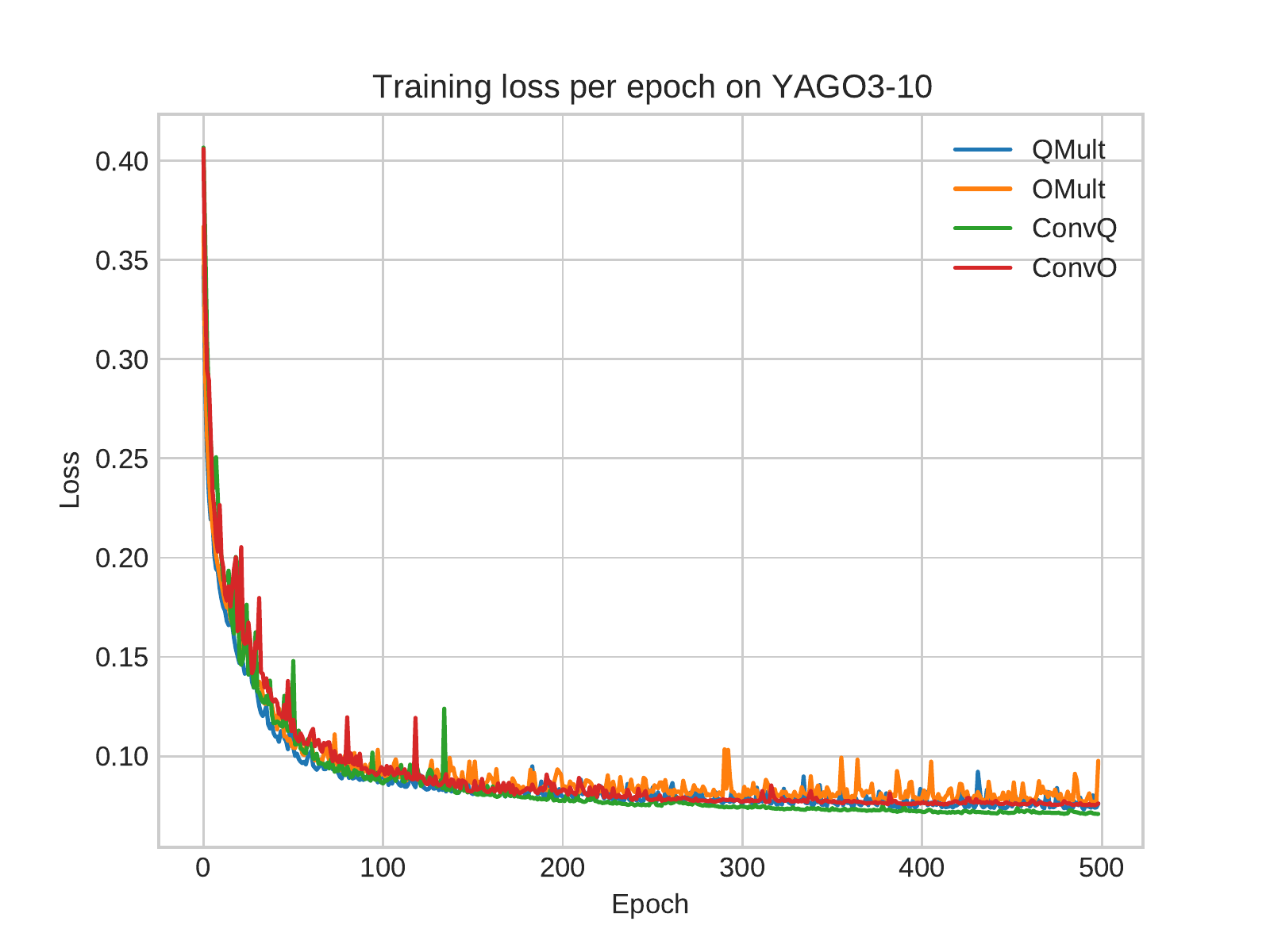}
    \caption{Convergence on the training set.}
    \label{fig:loss_on_yago3}
\end{figure}

\Cref{fig:loss_on_yago3} indicates that incurred binary cross entropy losses significantly decrease within the first 100 epochs. After the $300$th iteration, \approach and \approachocto appear to converge as losses do not fluctuate, whereas training losses of \baseapproach and \baseapproachocto continue fluctuating.
\begin{table*}[tbp]
\centering
\caption{Link prediction results on WN18 and FB15K. %Results are taken from~\cite{balavzevic2019tucker,nickel2015holographic,zhang2019quaternion}
}
\label{table:linkpred_on_wn18_fb15k}
% CD (16.04.2021) => results are double checked.
\footnotesize
%\scalebox{0.90}{%
\setlength{\tabcolsep}{4.1pt}
\begin{tabular}{@{}lcccccccccc@{}}
\toprule
               & \multicolumn{5}{c}{\textbf{WN18}}   & \multicolumn{5}{c}{\textbf{FB15K}} \\
               \cmidrule(lr){2-6} \cmidrule(l){7-11}
\textbf{Model}& Param. & MRR         &Hit@10       &Hit@3        &Hit@1       &Param.& MRR   & Hit@10 & Hit@3 & Hit@1 \\\midrule
TransE        & -& .495      & .943         &.888          &.113 & - &.463 & .749 & .578 & .297\\
TransR        & -& .605      & .940         &.876          &.335 & - &.346 & .582 & .404 & .218\\
ER-MLP        & -& .712      & .863         &.775          &.626 & - &.288 & .501 & .317 & .173\\
RESCAL        & -& .890      & .928         &.904          &.842 & - &.354 & .587 & .409 & .235\\
HolE          & -      & .938         &.949          &.945          &.930         &-     & .524 & .739  & .613 & .402 \\
SimplE        & -      & .942        &.947         &.944         &.939         &- & .727 & .838 & .773 & .660 \\
TorusE        & -      & .947         &.954          &.950          &.943         &- & .733 &.832& .771& .674 \\
RotatE        & -      & .947        &.961         &.953         &.938        &- & .699 & .872  & .788 &  .585 \\
QuatE         & -      & .949        &.960         &.954         &.941        &- &.770 & .878 & .821 & .700 \\
QuatE$^2$     & -      & .950        &.962         &.954         &.944        &- &\textbf{.833}& .900 & .859 & \textbf{.800} \\ 
\midrule
\baseapproach & 16.39M &\underline{.975}&\underline{.980}&\underline{.976}&\underline{.972}&7.05M &.755 & \underline{.896} & .819 & .668 \\
\baseapproachocto&16.39M&\underline{.975}&\textbf{.981}&\underline{.976}&\underline{.972}&7.05M &.748 & .889 & .813 & .660 \\
\approach & 21.51M      &\textbf{.976}   &\underline{.980}           &\textbf{.977}           & \textbf{.973}           &12.17M&\underline{.813} & \textbf{.923} &  \textbf{.868} & \underline{.743} \\
\approachocto & 21.51M  &\textbf{.976}   &\underline{.980}           &\textbf{.977}           & \textbf{.973}           &12.17M&.810 & \textbf{.923} & \underline{.865} & .739 \\\bottomrule
\end{tabular}%}
\end{table*}

\subsection{Link Prediction Results on Previous Benchmark Datasets}
\Cref{table:linkpred_on_wn18_fb15k} reports results on WN18 and FB15K showing that our approaches \approach and \approach outperform state-of-the-art approaches in 6 out of 8 metrics on the datasets.

\section{Discussion}
\label{sec:discussion}
% CD: Ready for the final submission
Our approaches often outperform many state-of-the-art approaches on all benchmark datasets. \baseapproach and \baseapproachocto outperform many state-of-the-art approaches including DistMult and ComplEx. These results indicate that scoring functions based on hypercomplex multiplications are more effective than scoring functions based on real and complex multiplications. This observation corroborates findings of~\cite{zhang2019quaternion}. \approachocto often perform slightly better than \approach on all datasets. Additionally, \baseapproach and \baseapproach perform particularly well on YAGO3-10. Figure 1 indicates that ConvO reaches a lower training error than QMult on YAGO3-10. This suggests that ConvO might have overfitted despite its higher expressiveness due to non-linearities and convolutions. In future work, we will design a regularization technique tailored to convolutions. % These results may stem from the fact that~\approach and \approachocto may benefit from initializing parameters with the correct variance as highlighted in~\citet{hanin2018start}.
Overall, superior performance of our models stems from (1) hypercomplex embeddings and (2) the inclusion of convolution operations. Our models are allowed to degrade into ComplEx or DistMult if necessary (see \Cref{sec:approach}). Inclusion of the convolution operation followed by an affine transformation permits finding a good ratio between expressiveness and the number of parameters.
\section{Conclusion}
\label{sec:conclusion}
% CD: Ready for the final submission
In this study, we presented effective compositions of convolution operations with hypercomplex multiplications in the quaternion and octonion algebras to address the link prediction problem. Experimental results showed that \baseapproach and \baseapproachocto performing hypercomplex multiplication on hypercomplex-valued embeddings of entities and relations are effective methods to tackle the link prediction problem. \approach and \approachocto forge
\baseapproach and \baseapproachocto with convolution operations followed by an affine transformation. By virtue of this novel composition, \approach and \approachocto facilitate finding a good ratio between expressiveness and the number of parameters. Experiments suggest that (1)~generalizing real- and complex-valued models such as DistMult and ComplEx to the hypercomplex space is beneficial, particularly for larger knowledge graphs, (2) the scaling effect of hypercomplex multiplication can be more effectively tackled with batch normalization than unit normalization, and (3) the application of ensembling can be used to further increase generalization performance. 

In future work, we plan to investigate the generalization of our approaches to temporal knowledge graphs and translation based models on hypercomplex vector spaces.
\section*{Acknowledgments}
This work has been supported by the German Federal Ministry of Education and Research (BMBF) within the project DAIKIRI under the grant no 01IS19085B and by the the German Federal Ministry for Economic Affairs and Energy (BMWi) within the project RAKI under the grant no 01MD19012B. We are grateful to Pamela Heidi Douglas for proofreading the manuscript.

\bibliography{references}
\end{document}